\documentclass[letterpaper, 10 pt, conference]{ieeeconf}
\IEEEoverridecommandlockouts
\usepackage[english]{babel}
\usepackage{amsmath}
\usepackage{amssymb}
\usepackage{booktabs}
\usepackage{caption}
\usepackage{subfigure}
\usepackage[dvips]{graphicx}
\usepackage{multirow}
\usepackage{amsfonts}
\usepackage{breakurl}
\usepackage{enumerate}
\usepackage{tabularx}
\usepackage{algorithm,algorithmic}
\usepackage{bm}
\usepackage{enumerate}
\usepackage{adjustbox}
\usepackage{xcolor}
\usepackage{siunitx}
\usepackage{booktabs}
\usepackage{mdframed}
\usepackage{adjustbox}
\usepackage{authblk}
\usepackage{color}
\usepackage{xurl}
\usepackage{cite}
\makeatletter
\let\NAT@parse\undefined
\makeatother
\usepackage{hyperref}
\usepackage{relsize}
\usepackage{float}
\usepackage{pifont}

\usepackage{fancyhdr}
\fancypagestyle{withfooter}{
  
  \fancyhead[L]{}
  \fancyhead[R]{}
  \fancyfoot[C]{\footnotesize Presented at the 2025 IEEE ICRA Workshop on Field Robotics}
}

\usepackage{stackengine,scalerel}

\usepackage{fontawesome5}

\DeclareMathOperator*{\argmin}{arg\,min}

\title{\LARGE \bf Reactive Collision Avoidance for Safe Agile Navigation}

\author{Alessandro Saviolo, Niko Picello, Jeffrey Mao, Rishabh Verma, and Giuseppe Loianno
\thanks{The authors are with the New York University, Brooklyn, NY 11201 USA {\tt\footnotesize \{as16054, np2965, jm7752, rv2378, loiannog\}@nyu.edu}.}
\thanks{This work was supported by the NSF CAREER Award 2145277, DARPA YFA Grant D22AP00156-00, Qualcomm Research, Nokia, NYU Wireless.}
}

\begin{document}

\thispagestyle{withfooter}
\pagestyle{withfooter}

\makeatletter
\g@addto@macro\@maketitle{
    \setcounter{figure}{0}
    \centering
    \includegraphics[width=\linewidth, trim=0 290 0 0, clip]{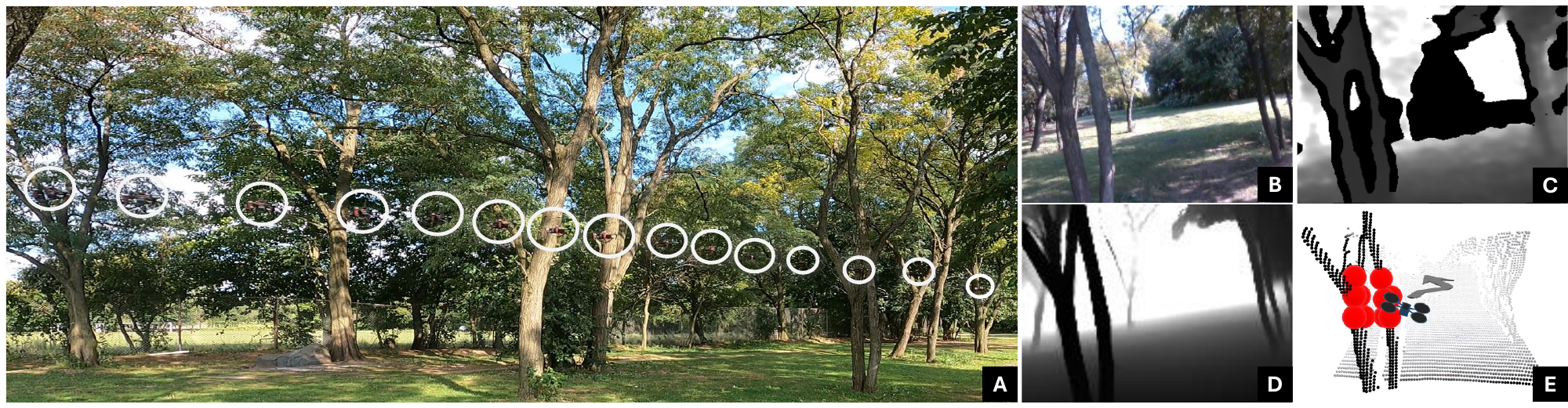}
    \captionof{figure}{
    \textbf{Safe and reactive collision avoidance for agile navigation in unstructured environments.} 
    The quadrotor navigates through a forest (A) using sensory inputs from an RGB-D camera (B, C), which are refined into a completed depth map (D) by a monocular depth estimation network. High-risk collision points are identified (red) and dynamically used as safety constraints within a nonlinear model predictive controller, enabling real-time reactive collision avoidance (E).
    \label{fig:initial_figure}}
}
\makeatother

\maketitle

\begin{abstract} 
Reactive collision avoidance is essential for agile robots navigating complex and dynamic environments, enabling real-time obstacle response.
However, this task is inherently challenging because it requires a tight integration of perception, planning, and control, which traditional methods often handle separately, resulting in compounded errors and delays.
This paper introduces a novel approach that unifies these tasks into a single reactive framework using solely onboard sensing and computing.
Our method combines nonlinear model predictive control with adaptive control barrier functions, directly linking perception-driven constraints to real-time planning and control. 
Constraints are determined by using a neural network to refine noisy RGB-D data, enhancing depth accuracy, and selecting points with the minimum time-to-collision to prioritize the most immediate threats.
To maintain a balance between safety and agility, a heuristic dynamically adjusts the optimization process, preventing overconstraints in real time.
Extensive experiments with an agile quadrotor demonstrate effective collision avoidance across diverse indoor and outdoor environments, without requiring environment-specific tuning or explicit mapping.
\end{abstract}

\section*{Supplementary Material}
\noindent \textbf{Video}: \url{https://youtu.be/pUiWym4NsvA}

\section{Introduction} \label{sec:introduction}
Reactive collision avoidance is essential for robots navigating dynamic and unstructured environments where pre-mapped data is unavailable or unreliable~\cite{huang2019collision}. 
This capability allows robots to perceive their surroundings and plan and execute collision-free maneuvers using real-time sensory observations, enabling them to adapt swiftly to unexpected changes such as moving obstacles or varying terrains. However, achieving reliable and efficient reactive collision avoidance remains a significant challenge due to the complexities of jointly solving perception, planning, and control~\cite{falanga2018pampc, falanga2019fast, song2023reaching, saviolo2024unifying, nagabandi2020deep, sanket2023ajna}.

A primary challenge lies in embedding raw perception data into control algorithms. 
Sensors like RGB-D cameras provide high-dimensional and noisy data, which are computationally expensive to process and often unreliable when directly used in control~\cite{grunnet2018best}. 
Many existing systems attempt to mitigate these issues by simplifying perception inputs or assuming ideal sensor performance~\cite{zhang2020optimization, lindqvist2020nonlinear, herbert2017fastrack, schulman2014motion, frasch2013auto, schoels2020nmpc, adajania2023amswarm}. Still, these simplifications often fall short in real-world scenarios where environmental conditions are unpredictable and sensor data is imperfect.

Jointly solving motion planning and control under perception constraints also presents substantial difficulties. 
Traditional methods often decouple all these tasks~\cite{droeschel2016multilayered, kanellakis2018towards, goricanec2023collision, lavalle2001randomized, kavraki1996probabilistic, ginesi2019dynamic, heng2011autonomous, hart1968formal, zhao2024learning, marcucci2023motion, meijer2024pushing, tordesillas2019real}, resulting in inefficiencies and suboptimal performance in dynamic environments.
In contrast, Nonlinear Model Predictive Control (NMPC) addresses these limitations by integrating planning and control into a unified constrained optimization framework. This integration allows NMPC to directly embed safety constraints, such as obstacle avoidance and actuation limits, within its optimization process, making it particularly effective for reactive collision avoidance~\cite{lin2020robust, jacquet2024n, krinnermpcc}.

Control Barrier Functions (CBFs) have recently emerged as a widely adopted method for enforcing safety constraints in obstacle avoidance~\cite{wang2024dual, zeng2021safety, son2019safety, depoint, cosner2022self, tong2023enforcing, liu2024flexible, khan2020barrier, dai2024sailing, niu2021vision, tayal2024collision, jang2024safe, dai2023safe}. 
CBFs provide a formal mathematical framework that defines boundaries to prevent a system from entering unsafe states. 
By establishing a barrier between safe and unsafe regions of the state space, CBFs guide control inputs to maintain safety while achieving the desired task~\cite{ames2019control}. 
However, despite their theoretical appeal, state-of-the-art approaches often oversimplify CBFs as safety filters~\cite{depoint}, rely on restrictive assumptions of static obstacles~\cite{cosner2022self, wang2024dual, depoint, zeng2021safety, son2019safety}, and face computational challenges~\cite{tong2023enforcing, liu2024flexible, khan2020barrier}, limiting their applicability to real-world settings.

To address these challenges, we present a novel framework that integrates NMPC with dynamically adjusting high-order CBFs to jointly solve the planning and control tasks under real-time perception constraints. Our contributions are: (i) We propose a unified approach that adapts to unknown environments without relying on assumptions about obstacle characteristics or sensor performance. (ii) Our perception pipeline utilizes a monocular depth estimation neural network to enhance noisy RGB-D sensor data and identify high-risk collision points based on minimal time-to-collision. (iii) We introduce dynamically adjusting high-order CBFs that update at every NMPC iteration, allowing the system to continuously adapt its safety constraints. (iv) A heuristic is implemented to dynamically relax the optimization process based on real-time collision risk, balancing safety and performance. Extensive experiments on an agile quadrotor platform demonstrate the efficiency and effectiveness of the proposed framework in both indoor and outdoor environments, proving its real-time feasibility for applications where reactive collision avoidance is critical to ensure safety (Figure~\ref{fig:initial_figure}).

\section{Related Work} \label{sec:related_works}
\textbf{Perception and Environment Sensing.}
Accurate environmental perception is crucial for reactive collision avoidance in autonomous robots, particularly when pre-mapped data is unavailable or unreliable. Traditional sensors like LiDAR and sonar have been widely used due to their precision in depth measurement~\cite{harms2024neural, ahn2022model, ebadi2022present, gadde2021fast, keyumarsi2023lidar, pantic2023obstacle}. 
Still, their bulky size, high weight, and significant energy consumption make them unsuitable for small, agile platforms such as quadrotors~\cite{park2020collision}. 
RGB-D cameras are frequently adopted for onboard sensing as they offer a lightweight and cost-effective alternative with a wide field of view.
However, these sensors often produce high-dimensional, noisy, and incomplete data due to inherent sensor limitations, adverse environmental conditions, and the variable reflectivity of different surfaces~\cite{grunnet2018best}. 
These data imperfections pose significant challenges when using perception data for motion planning and control directly.

To mitigate these challenges, recent research has increasingly focused on deep learning methods for enhancing depth perception. Techniques such as monocular depth estimation utilize visual cues including shading, texture gradients, and object boundaries, achieving strong generalization across diverse environments~\cite{birkl2023midas, ranftl2021vision, wang2024dust3r}. Nonetheless, these methods predominantly estimate relative depth and often struggle to accurately infer absolute depth, which is essential for precise obstacle avoidance~\cite{piccinelli2024unidepth}. Efforts to improve absolute depth accuracy through scale correction and environment-specific priors have been explored~\cite{wofk2019fastdepth, bhat2023zoedepth, hu2024metric3d, yang2024depth2}, but these solutions often come at the cost of increased computational complexity and reduced adaptability to new environments.

\textbf{Collision-Free Planning and Control.} 
Traditional collision avoidance methods often separate planning and control, where the planner generates a collision-free path, and the controller is tasked to follow it. Common methods include potential fields~\cite{droeschel2016multilayered, kanellakis2018towards, goricanec2023collision}, sampling-based~\cite{lavalle2001randomized, kavraki1996probabilistic}, motion primitives~\cite{ginesi2019dynamic}, and graph-search methods~\cite{heng2011autonomous, hart1968formal}. This separation has limitations, as it ignores disturbances during the control phase and can result in paths the controller cannot feasibly execute. Additionally, decoupling leads to delays, reducing responsiveness in rapidly changing environments.

Optimization-based methods such as NMPC address these limitations by merging planning and control into a single process. NMPC predicts future states and optimizes control inputs in real-time, embedding constraints directly into the control loop, offering a well-suited framework for obstacle avoidance~\cite{zhang2020optimization, lindqvist2020nonlinear, herbert2017fastrack, schulman2014motion, frasch2013auto, schoels2020nmpc}. 
Among the various safety constraints that can be integrated into NMPC, CBFs are particularly effective due to their formal mathematical guarantees of safety~\cite{depoint, cosner2022self, tong2023enforcing}. High-order CBFs further enhance adaptability by allowing dynamic adjustment of safety boundaries in response to changes in robot and obstacle dynamics~\cite{ames2019control}. This integrated approach, using CBFs or other safety constraints within NMPC, has demonstrated success in applications such as robotic manipulation~\cite{liu2024flexible}, coordinated quadrotor swarms~\cite{wang2024dual}, and high-speed autonomous vehicles~\cite{zeng2021safety, son2019safety}.

Despite these advances, real-world implementation remains challenging, as many studies assume perfect sensing and known obstacle locations. 
Extending NMPC with safety constraints like CBFs to handle unknown, dynamic environments is essential. Integrating perception, planning, and control in a unified framework enables systems to adapt control strategies in real-time, maintaining safety without compromising performance, even with imperfect data.

Recent research has investigated learning-based methods for obstacle avoidance, focusing on approaches that either learn end-to-end policies~\cite{kulkarni2024reinforcement, hatch2021obstacle, lu2023lpnet, brilli2023monocular} or optimize specific components within the NMPC framework~\cite{jacquet2024n}. Despite their potential, these methods often struggle with challenges such as interpretability, safety compliance, high data requirements, and computational overhead, which limit their real-time applicability. Integrating learning-based techniques with NMPC and robust safety constraints offers a promising path to improve adaptability and performance, particularly in dynamic and unpredictable environments.

\begin{figure*}[t]
    \centering
    \includegraphics[width=\linewidth, trim=0 340 0 0, clip]{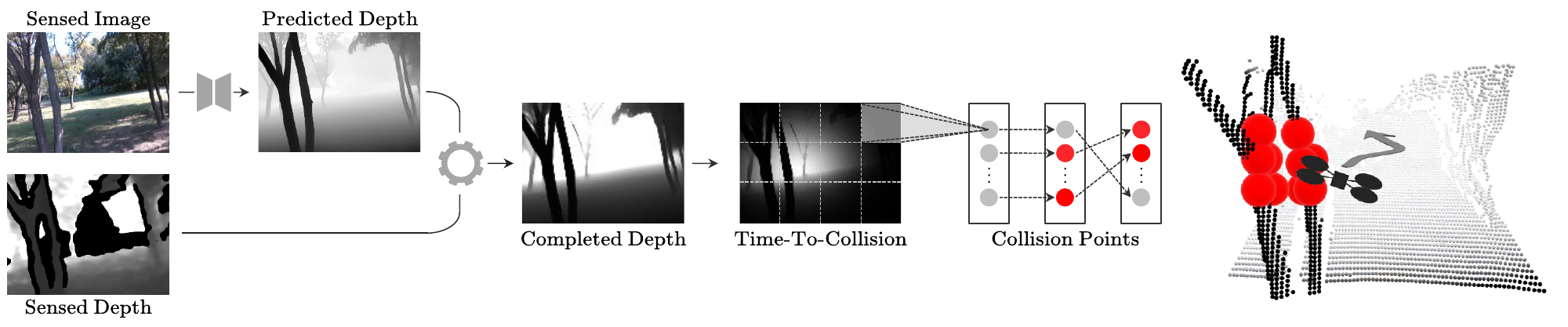}
    \caption{ \textbf{The proposed reactive collision avoidance framework.} The perception pipeline completes sensory data by combining the sensed depth map with a predicted depth map generated from a monocular depth estimation neural network. This map is processed to compute a time-to-collision map, identifying high-risk collision points based on the robot’s velocity. The points with the highest risk of collision (highlighted in red) are identified through pooling and sorting. These high-risk points dynamically define safety constraints within the NMPC framework through CBFs, enabling safe collision-free navigation.}
    \label{fig:method}
\end{figure*}

\section{Proposed Methodology} \label{sec:methodology}
The goal is to enable an agile aerial vehicle with an RGB-D sensor to navigate safely from start to goal in unknown, dynamic environments. We do not assume prior knowledge of the environment, including obstacle shape or quantity. The robot must perceive its surroundings in real time and make decisions to avoid collisions while reaching its goal.

Figure~\ref{fig:method} illustrates the proposed framework. 
The perception module processes RGB-D data to generate a depth map, which is then converted into a time-to-collision map to identify the $K$ highest-risk collision points. These points are integrated into the NMPC optimization via CBFs, allowing the controller to enforce safety constraints during flight.

\subsection{Depth Completion Using Monocular Depth Estimation}
RGB-D sensors provide RGB images $\mathbf{I}^{t} \in \mathbb{R}^{H \times W \times 3}$ and absolute depth maps $\mathbf{D}^{t}_{\text{abs}} \in \mathbb{R}^{H \times W}$ at each time $t$. However, these depth maps often suffer from noise, incompleteness, and inaccuracies due to sensor limitations.

To address these limitations and extend the perception range beyond the baseline of the stereo camera, we employ a monocular depth estimation network $\mathcal{N}_{\text{mde}}(\mathbf{I}^{t})$ that generates a relative depth map $\mathbf{D}^{t}_{\text{rel}} \in \mathbb{R}^{H \times W}$. The observed depth map is refined by aligning it with the predicted depth using a second-order polynomial fitted to valid pixels identified by a binary mask $\mathbf{M}^{t} \in {0, 1}^{H \times W}$
\begin{equation}
    \min_{a, b, c} \sum_{(i,j) \in \mathcal{M}} \left[ \mathbf{D}^{t}_{\text{abs}}(i,j) - \left(a \mathbf{D}^{t}_{\text{rel}}(i,j)^2 + b \mathbf{D}^{t}_{\text{rel}}(i,j) + c\right) \right]^2,
\end{equation}
where $\mathcal{M}$ denotes the set of valid pixel indices defined by the mask. 
Our empirical analysis demonstrates that the second-order polynomial provides the most accurate fit for capturing this relationship (Section~\ref{subsec:analysis_poly_order}).

The fitted polynomial coefficients $(a, b, c)$ allow us to compute the completed depth map
\begin{equation}
    \mathbf{D}^{t}_{\text{com}} = a \cdot (\mathbf{D}^{t}_{\text{rel}})^2 + b \cdot \mathbf{D}^{t}_{\text{rel}} + c \cdot \mathbf{1},
\end{equation}
which minimizes noise, fills missing values, and corrects inaccuracies, providing a reliable environment representation.

\subsection{Fast Identification of High-Risk Collision Points}
The completed depth map $\mathbf{D}^{t}_{\text{com}}$ is used to compute the time-to-collision (TTC) map $\mathbf{T}^{t} \in \mathbb{R}^{H \times W}$, which estimates collision likelihood based on the robot's velocity vector $\mathbf{v}^{t}$. 
The TTC map is calculated as
\begin{equation}
    \mathbf{T}^{t} = \frac{\mathbf{D}^{t}_{\text{com}}}{\|\mathbf{v}^{t} \cdot \mathbf{R}^{t}\|},
\end{equation}
where $\mathbf{R}^{t} \in \mathbb{R}^{H \times W \times 3}$ contains the 3D ray directions from the camera through each coordinate.

TTC represents the relative distance to collision coordinates in temporal units without requiring camera calibration or specific knowledge of the environment's structure, obstacle size, or shape~\cite{alenya2009comparison}. 
This approach allows fast computation, providing the basis for quick reactive avoidance maneuvers.
Despite its efficiency, the TTC map is dense, including $H \times W$ coordinates (potentially hundreds of thousands in standard camera image sizes). 

To retain only the most critical coordinates, the TTC map is partitioned into non-overlapping grid cells $\mathcal{P}_{u,v} = \{(i, j) \mid uP \leq i < uP + P, \, vP \leq j < vP + P\}$ of size $P \times P$. 
For each cell, a min-pooling operator selects the coordinate corresponding to the smallest TTC value, effectively filtering out redundant points and retaining the most critical ones that are spatially distinct rather than overly clustered.
This process forms a set of high-risk coordinates
\begin{equation}
    \mathcal{S}^{t}_{\text{ttc}} = 
    \left\{
        (i, j) \in \mathbf{T}^{t} \mid (i, j) = \argmin_{(i, j) \in \mathcal{P}_{u,v}} \mathbf{T}^{t}(i, j), \; \forall \, u, v 
    \right\}.
\end{equation}

Next, these coordinates are filtered to retain only those representing real collision threats by comparing each pooled coordinate against the robot's projected dimensions at the corresponding depth
\begin{equation}
    \mathcal{S}^{t}_{\text{filt}} = 
    \left\{ 
        (i, j) \in \mathcal{S}^{t}_{\text{ttc}} \; \middle| \;
        \begin{aligned}
            &\left| i - c_x \right| \leq Q_x f_x / [2\mathbf{D}^{t}_{\text{com}}(i, j)], \\
            &\left| j - c_y \right| \leq Q_y f_y / [2\mathbf{D}^{t}_{\text{com}}(i, j)]
        \end{aligned}
    \right\},
\end{equation}
where $(Q_x, Q_y)$ are the robot's dimensions, $(f_x, f_y)$ and $(c_x, c_y)$ are the camera’s focal lengths and principal point.

Finally, the top-$K$ coordinates with the lowest TTC values are selected from the refined set $\mathcal{S}^{t}_{\text{filt}}$, prioritizing the highest-risk coordinates for reactive control
\begin{equation}
    \mathcal{S}^{t}_{\text{top-}K} = 
    \argmin_{\substack{\mathcal{S'}^{t}_{\text{filt}} \subseteq \mathcal{S}^{t}_{\text{filt}}, |\mathcal{S'}^{t}_{\text{filt}}| = K}} 
    \sum_{(i, j) \in \mathcal{S'}^{t}_{\text{filt}}} 
    \mathbf{T}^{t}(i, j),
\end{equation}
where $\mathcal{S'}^{t}_{\text{filt}}$ is any subset of $\mathcal{S}^{t}_{\text{filt}}$ containing $K$ coordinates.
The pixel coordinates are then transformed into 3D points in the camera frame $\mathcal{C}$
\begin{equation}
    \mathcal{S}^{t}_{\mathcal{C}} = \left\{ 
    \left( 
    \frac{(i - c_x) z}{f_x}, \frac{(j - c_y) z}{f_y}, z 
    \right) 
    \; \middle| \; 
    \begin{aligned}
        &z = \mathbf{D}^{t}_{\text{com}}(i, j), \\
        &(i, j) \in \mathcal{S}^{t}_{\text{top-}K} 
    \end{aligned} 
    \right\}.
\end{equation}

These 3D points are finally converted from the camera frame $\mathcal{C}$ to the body $\mathcal{B}$ and world $\mathcal{W}$ frames:
%
\begin{equation}
    \mathcal{S}^{t}_{\text{W}} = 
    \left\{ 
    \mathbf{R}_{\mathcal{WB}} \cdot \left(\mathbf{R}_{\mathcal{BC}} \cdot \mathbf{s}_{\mathcal{C}} + \mathbf{t}_{\mathcal{BC}}\right) + \mathbf{t}_{\mathcal{WB}} 
    \; \middle| \; \mathbf{s}_{\mathcal{C}} \in \mathcal{S}^{t}_{\mathcal{C}} 
    \right\},
\end{equation}
where $\mathbf{R}_{\mathcal{BC}}$ and $\mathbf{t}_{\mathcal{BC}}$ are the rotation matrix and translation vector from the camera frame to the body frame, $\mathbf{R}_{\mathcal{WB}}$ and $\mathbf{t}_{\mathcal{WB}}$ are the rotation matrix and translation vector from the body frame to the world frame.

\subsection{Motion Planning and Control with Safety Constraints}
The NMPC framework computes optimal control inputs that minimize trajectory deviations while ensuring collision avoidance.
The state vector is defined as
\begin{equation*}
    \mathbf{x} = 
    \begin{bmatrix}
        \mathbf{p}^\top_\mathcal{W} & 
        \mathbf{v}^\top_\mathcal{W} & 
        \mathbf{R}^\top_\mathcal{BW} & 
        \boldsymbol{\omega}^\top_\mathcal{B} 
    \end{bmatrix}^\top,
\end{equation*}
where $\mathbf{p}$ is the system’s position in $\mathcal{W}$, $\mathbf{v}$ the linear velocity in $\mathcal{W}$, $\mathbf{R}$ the orientation from $\mathcal{B}$ to $\mathcal{W}$, and $\boldsymbol{\omega}$ the angular velocity in $\mathcal{B}$. 
This state forms a flexible representation for general aerial robots.
The control input vector, denoted by $\mathbf{u}$, is used to influence the system's dynamics $\mathbf{x}^{t+1} = f(\mathbf{x}^{t}, \mathbf{u}^{t})$.

To ensure safety, we employ second-order CBFs for each high-risk point $\mathbf{q}^{t}_{\mathcal{W}} \in \mathcal{S}^{t}_{\mathcal{W}}$ identified from perception data. These CBFs constrain the robot to maintain a safe distance from the obstacles. Each safety constraint is defined as
\begin{equation}
    h_k(\mathbf{x}^{t}) = \|\mathbf{q}^{t}_{\mathcal{W}} - \mathbf{p}^{t}_{\mathcal{W}}\|^2 - r_{\text{safe}}^2,
\end{equation}
where $r_{\text{safe}} = \max \{Q_x, Q_y\} + \psi$ combines the maximum dimension of the robot and a tolerance $\psi$ to accommodate estimation uncertainties. This formulation ensures that the safety set remains forward-invariant, providing a boundary that responds dynamically to both position and velocity.

The second-order CBF constraint is expressed as
\begin{equation}
    \ddot{h}_k(\mathbf{x}^{t}, \mathbf{u}^{t}) + 2\lambda\dot{h}_k(\mathbf{x}^{t}) + \lambda^2 h_k(\mathbf{x}^{t}) \geq 0,
\end{equation}
where the parameter $\lambda > 0$ modulates the aggressiveness of the constraint enforcement, $\dot{h}_k(\mathbf{x}^{t})$ captures the approach rate to the safety boundary
\begin{equation}
    \dot{h}_k(\mathbf{x}^{t}) = 2(\mathbf{q}^{t}_{\mathcal{W}} - \mathbf{p}^{t}_{\mathcal{W}})^\top \mathbf{v}^{t}_{\mathcal{W}},
\end{equation}
and $\ddot{h}_k(\mathbf{x}^{t}, \mathbf{u}^{t})$ reflects the influence of control inputs on the acceleration towards the boundary
\begin{equation}
    \ddot{h}_k(\mathbf{x}^{t}, \mathbf{u}^{t}) = 2 \mathbf{v}^{t\top}_{\mathcal{W}} \mathbf{v}^{t}_{\mathcal{W}} + 2(\mathbf{q}^{t}_{\mathcal{W}} - \mathbf{p}^{t}_{\mathcal{W}})^\top \dot{\mathbf{v}}_{\mathcal{W}}(\mathbf{u}),
\end{equation}
with $\dot{\mathbf{v}}_{\mathcal{W}}(\mathbf{u})$ as the robot’s acceleration driven by $\mathbf{u}$.

Integrating these CBF constraints into the NMPC scheme, the optimization problem is formulated as follows
\begin{equation}
    \min_{\substack{
    \mathbf{x}^{t},\dots,\mathbf{x}^{t+N}\\
    \mathbf{u}^{t},\dots,\mathbf{u}^{t+N-1}}}
    \sum_{j=0}^{N} \Tilde{\mathbf{x}}^{t+j\top} \mathbf{Q}_{\mathbf{x}}^{t} \Tilde{\mathbf{x}}^{t+j} + \sum_{j=0}^{N-1} \Tilde{\mathbf{u}}^{t+j\top} \mathbf{Q}_{\mathbf{u}}^{t} \Tilde{\mathbf{u}}^{t+j},
\end{equation}
where $\Tilde{\mathbf{x}}^{t+j} = \mathbf{x}_{\mathrm{des}}^{t+j} - \mathbf{x}^{t+j}$ and $\Tilde{\mathbf{u}}^{t+j} = \mathbf{u}_{\mathrm{des}}^{t+j} - \mathbf{u}^{t+j}$ represent the deviations between desired and actual states and inputs. 
The constraints applied $\forall j \in [0, N]$ and $\forall k \in [0, K)$ are
\begin{subequations}
    \begin{align}
        &\mathbf{x}^{t} = \hat{\mathbf{x}}^{t}, \\
        &\mathbf{x}^{t+1+j} = f(\mathbf{x}^{t+j}, \mathbf{u}^{t+j}), \\
        &\mathbf{x}_{\min} \leq \mathbf{x}^{t+j} \leq \mathbf{x}_{\max}, \\
        &\mathbf{u}_{\min} \leq \mathbf{u}^{t+j} \leq \mathbf{u}_{\max}, \\
        &\ddot{h}_k(\mathbf{x}^{t+j}, \mathbf{u}^{t+j}) + 2 \lambda \dot{h}_k(\mathbf{x}^{t+j}) +\lambda^2 h_k(\mathbf{x}^{t+j}) \geq 0,
    \end{align}
\end{subequations}
where $\mathbf{Q}^{t}_{\mathbf{x}}$ and $\mathbf{Q}^{t}_{\mathbf{u}}$ are positive semi-definite weight matrices, and $\hat{\mathbf{x}}^t$ is the current state estimate.
Importantly, the CBFs are dynamically adjusted at each NMPC iteration, allowing the robot to adapt its trajectory to avoid obstacles and maintain safety even when there are drifts in state estimation.

When $K$ is large, the optimization problem may become overconstrained, forcing the NMPC to incur high costs to maintain reference tracking while satisfying all safety constraints. To prioritize safety, it is crucial to relax the tracking requirements, allowing a greater margin of error in trajectory performance. To address this challenge, we propose a heuristic that dynamically adjusts the NMPC cost function.
First, we identify the point with the lowest TTC 
\begin{equation}
    (i^*, j^*) = \argmin_{(i, j) \in \mathcal{S}^{t}_{\text{ttc}}} \mathbf{T}^{t}(i, j).
\end{equation}

Next, the TTC is normalized using a sigmoid function to modulate its impact on the control objectives
\begin{equation}
    \nu^{t} = (1 + \exp \{-a(\mathbf{T}^{t}(i^*, j^*) - b) \})^{-1},
\end{equation}
where the parameters $a$ and $b$ adjust the sigmoid’s sensitivity, balancing collision avoidance with control performance.

Finally, we adjust the NMPC weight matrices as follows
\begin{equation}
    \begin{aligned}
        \mathbf{Q}_{\mathbf{x}}^{t} &= \bar{\mathbf{Q}}_{\mathbf{x}} \cdot \nu^{t} ~~~ \text{and} ~~~ \mathbf{Q}_{\mathbf{u}}^{t} &= \bar{\mathbf{Q}}_{\mathbf{u}},
    \end{aligned}
\end{equation}
where $\bar{\mathbf{Q}}_{\mathbf{x}}$ and $\bar{\mathbf{Q}}_{\mathbf{u}}$ are the traditional fixed weight matrices in NMPC tuned for trajectory tracking~\cite{saviolo2021autotune}. 


\begin{figure*}[t]
    \centering
    \includegraphics[width=\linewidth, trim=0 800 0 0, clip]{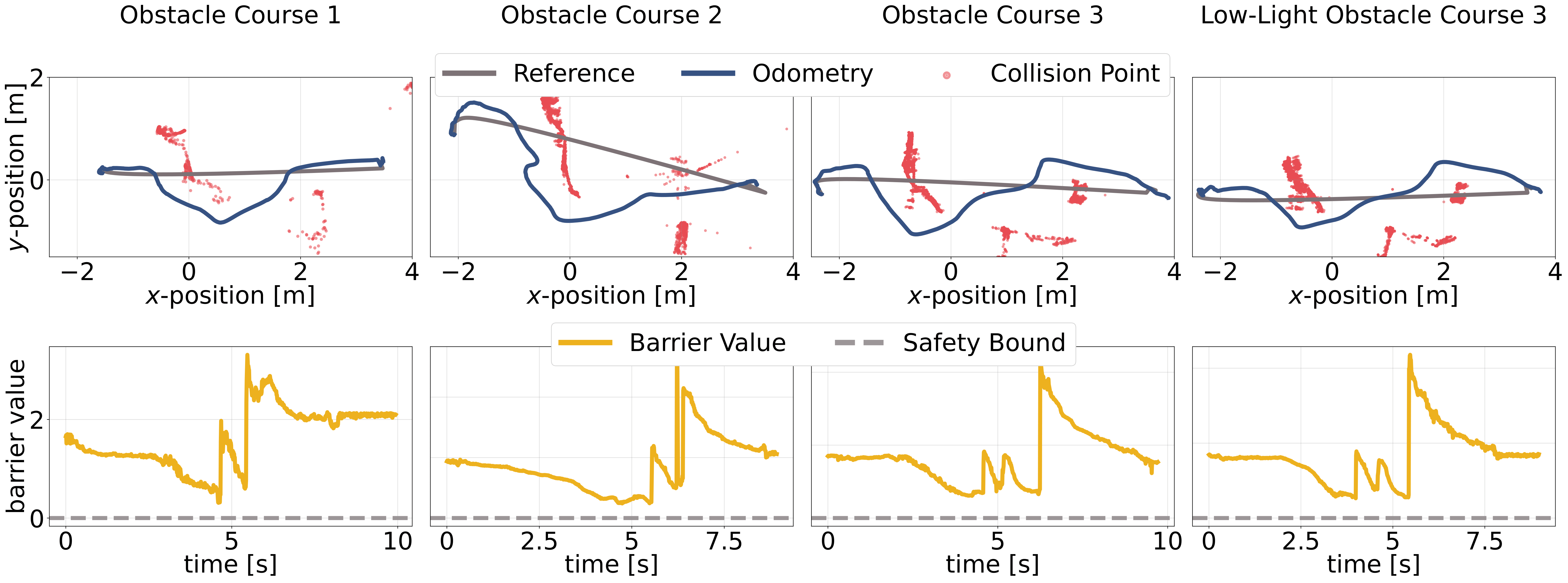}\\
    ~~
    \includegraphics[width=0.235\linewidth, trim=50 0 150 100, clip]{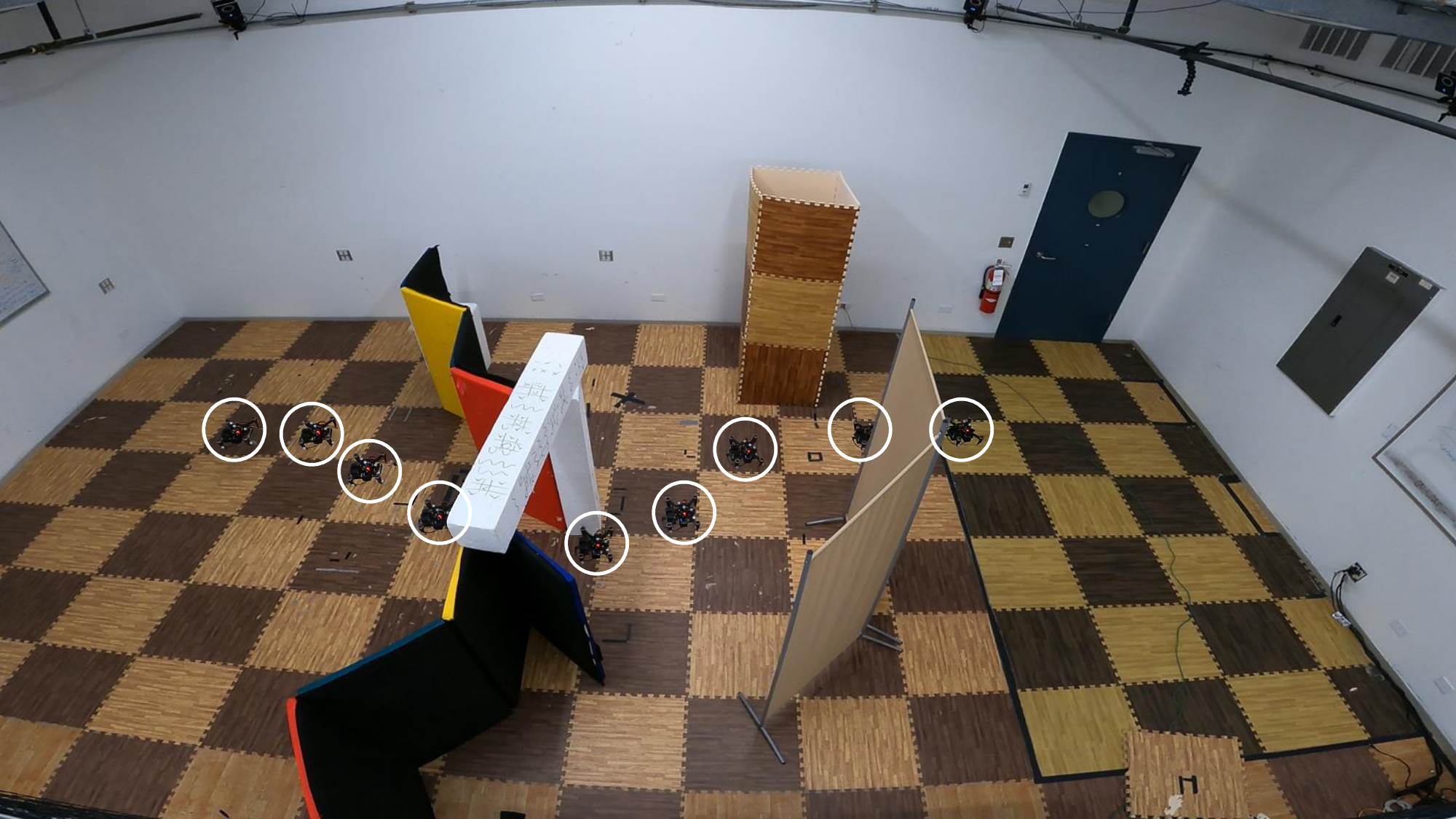}
    \includegraphics[width=0.235\linewidth, trim=100 50 100 50, clip]{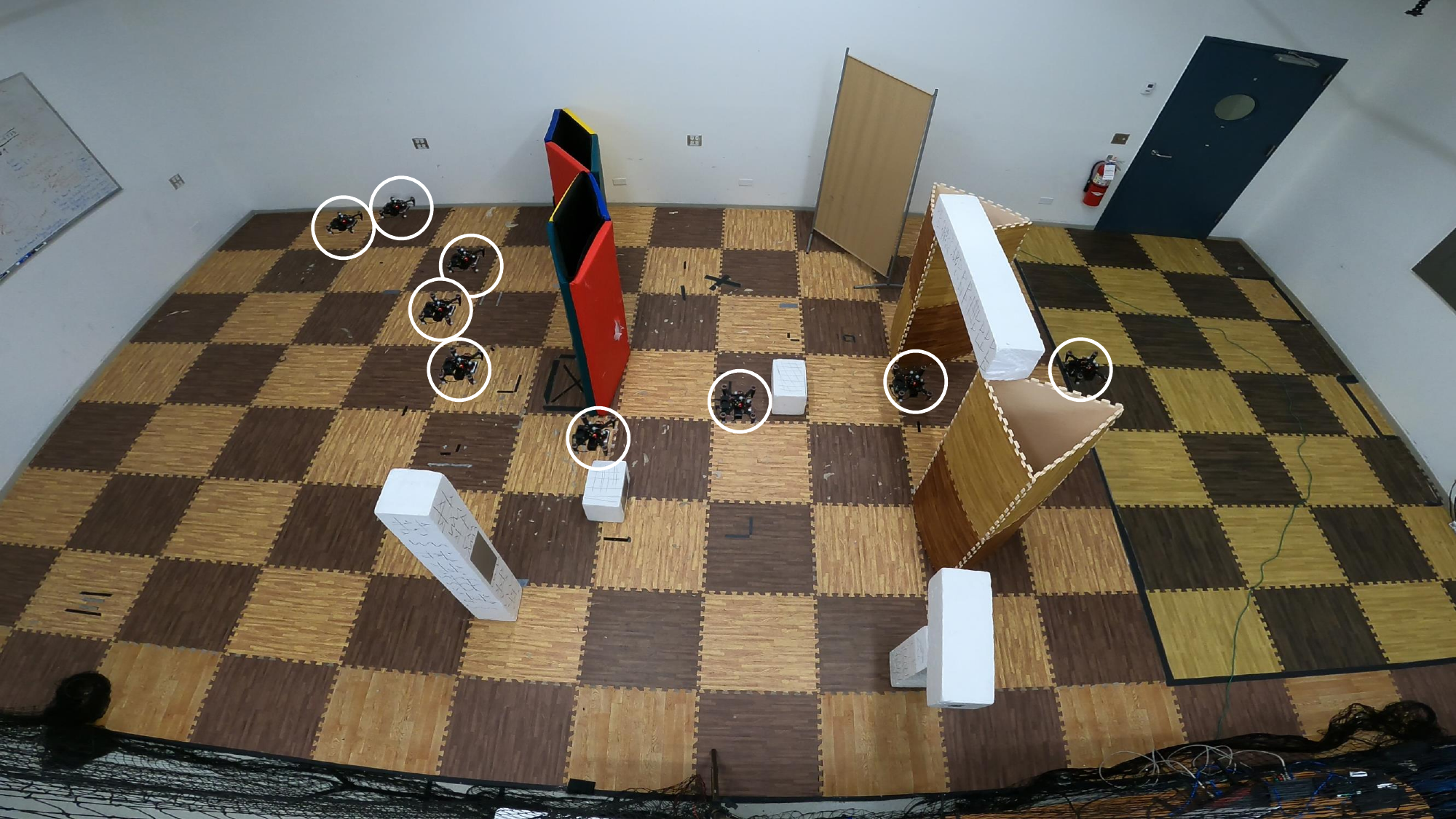}
    \includegraphics[width=0.235\linewidth, trim=50 50 150 50, clip]{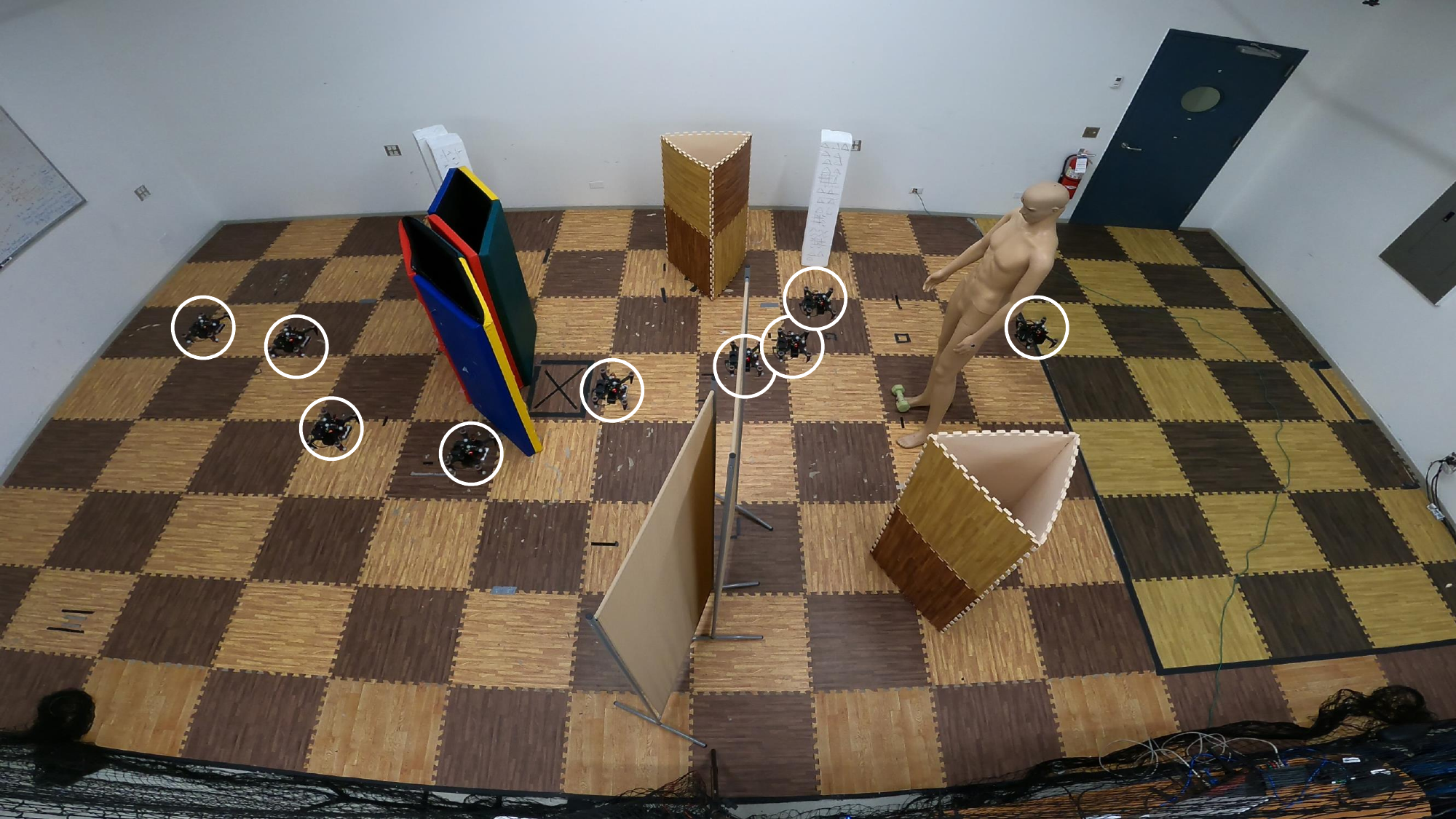}
    \includegraphics[width=0.235\linewidth, trim=50 50 150 50, clip]{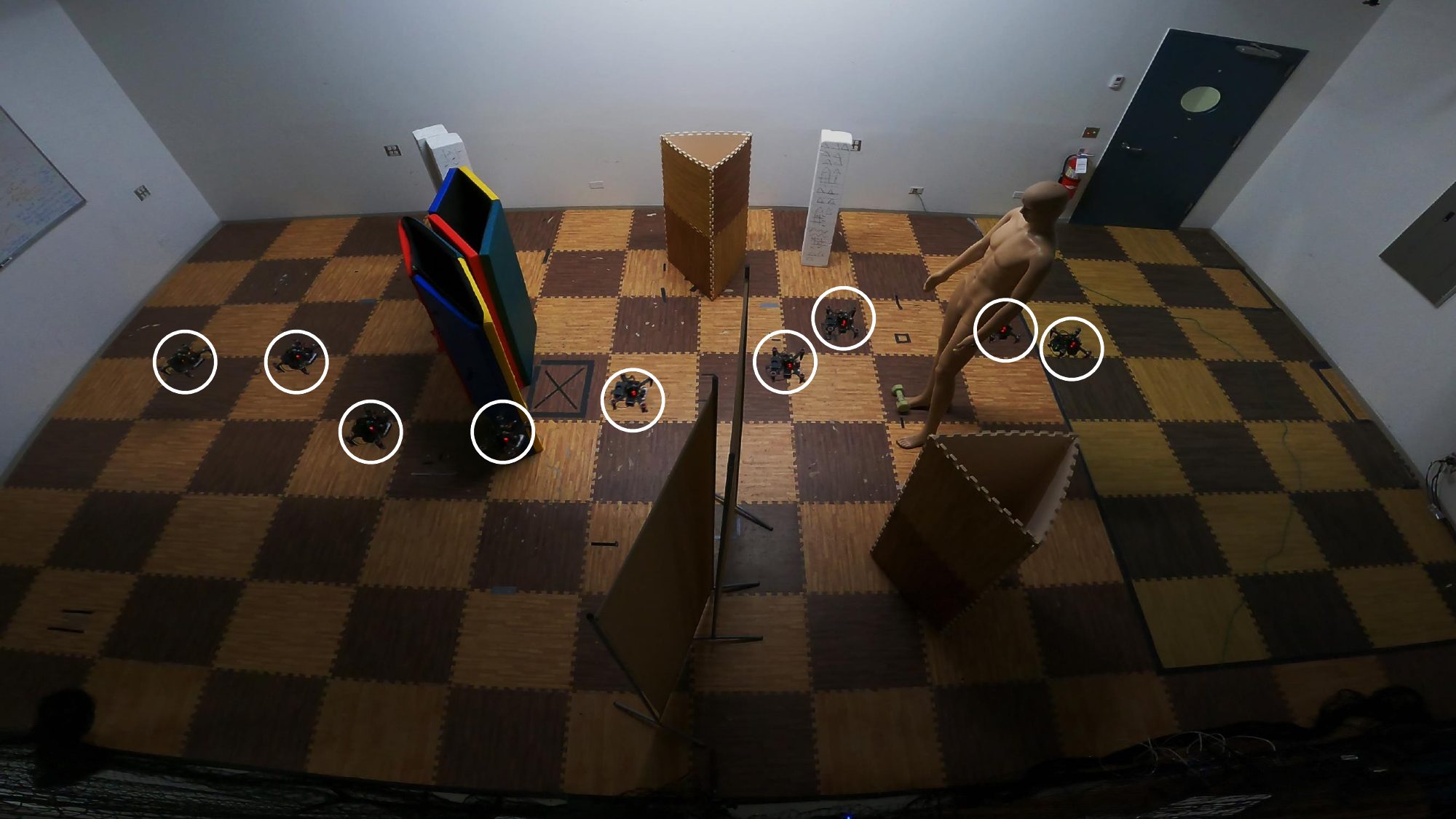}\\
    \vspace{1em}
    \includegraphics[width=\linewidth, trim=0 0 0 78, clip]{images/indoor-1.png}
    \caption{
        \textbf{Experimental results demonstrating reactive collision avoidance in unknown and cluttered indoor environments.} 
        \emph{(Top row)}: A composite image showing the quadrotor navigating various indoor settings with unknown obstacles. 
        \emph{(Middle row)}: Top-down views illustrating the reference paths and the actual collision-free paths tracked by the quadrotor. High-risk points identified during navigation are highlighted, showing the system's responsiveness to potential collisions. 
        \emph{(Bottom row)}: Graphs of the minimum values of collision-avoidance CBFs over time. The non-negative values indicate successful enforcement of safety constraints, ensuring obstacle avoidance throughout the experiment.
    }
    \label{fig:collision_avoidance_indoor}
    \vspace{-10pt}
\end{figure*}

\section{Experimental Setup and Implementation} \label{sec:exp_setup}
\textbf{System.}
Our quadrotor weighs $1.2~\si{kg}$ with a thrust-to-weight ratio of $4$ and a motor span of $25~\si{cm}$. Powered by NVIDIA Jetson Orin NX, it processes tasks onboard. A forward-facing Intel RealSense D435i captures $320 \times 240$ frames. The NMPC module runs at $200~\si{Hz}$ per state estimation, and CBFs update at $60~\si{Hz}$ after each RGB-D frame.

\textbf{Facility.} 
Experiments took place in the ARPL indoor lab at New York University and in outdoor fields in New York State. The indoor space ($10 \times 6 \times 3~\si{m^3}$) featured controlled lighting and static obstacles like poles and bars. Outdoor tests included natural (trees) and artificial (soccer goals) obstacles under varying weather and light conditions to test robustness.

\textbf{Localization.}
Indoor localization used a Vicon motion capture system, providing sub-millimeter accuracy at $200~\si{Hz}$. Outdoor experiments relied on a GPS u-Blox Neo-M9N module fused with IMU data via an Extended Kalman Filter (EKF), providing position updates at $100~\si{Hz}$.

\textbf{Perception.} 
Depth data was acquired with the Intel RealSense D435i using the HighAccuracy setting~\cite{grunnet2018best}, though the sensor produced incomplete data around edges and reflective surfaces. Monocular depth estimation used DepthAnythingV2~\cite{yang2024depth2} with a Vision Transformer (ViT-S) backbone, optimized for NVIDIA Jetson using TensorRT, reducing inference time to $19~\si{ms}/\text{frame}$. Collision points were identified using a pooling kernel of size $P=17$, downsampling depth data to highlight the top $K=10$ high-risk coordinates. These were projected into 3D, transformed into the world frame, and modeled as CBF constraints with a safety gain $\lambda=2$ (selected empirically following~\cite{depoint}).

\textbf{Planning and Control.}
The NMPC framework uses $10$ shooting steps over a $1~\si{s}$ prediction horizon, solved via sequential quadratic programming with \texttt{acados} \cite{acados}, leveraging a Gauss-Newton Hessian approximation and Levenberg-Marquardt regularization. The control input $\mathbf{u}$ includes four motor thrusts, governed by a dynamic model from previous works \cite{saviolo2022pitcn, saviolo2023active, saviolo2023learning}. A min-jerk trajectory planner ensures smooth transitions, while differential flatness principles derive the desired control actions. Carefully tuned state and input weights enable precise trajectory tracking. The NMPC runs efficiently with $10$ active CBF constraints, validated experimentally, achieving an average computation time of $2~\si{ms}$ per iteration.

\section{Results and Performance Evaluation} \label{sec:exp_results}
We evaluate the proposed approach through experiments in cluttered indoor and outdoor environments. In indoor tests, the system is exposed to diverse obstacle configurations, including varying shapes, sizes, and densities, demonstrating its ability to generalize and maintain real-time performance with adjusting CBFs, ensuring safe navigation without prior knowledge of the environment. Outdoor experiments focus on agile reactive collision avoidance to maneuver the robot through trees and around the poles of a soccer goal. For further visual insights, please refer to the video.

Additionally, we analyze the impact of polynomial order on depth completion and the computational complexity across camera resolutions and constraint numbers.

\subsection{Real-Time Performance in Cluttered Indoor Environments}
We evaluate the system’s performance in various cluttered indoor settings with diverse obstacle configurations, including styrofoam cuboids, mats with contrasting and ground-matching textures, room dividers, and mannequins. 
To further assess the robustness of the perception pipeline, we conduct tests also under low-light conditions.
Detailed results and environment setups are depicted in Figure~\ref{fig:collision_avoidance_indoor}.

The quadrotor successfully navigates these complex scenarios without pre-tuning or prior adjustments to the perception, planning, or control modules, demonstrating the effectiveness and adaptability of the integrated NMPC with dynamically adjusting CBFs. 
Throughout all tests, CBF values consistently stayed above the safety thresholds, confirming the system's ability to maintain safety constraints even in dense and low-visibility environments. 

\subsection{Agile Safe Navigation in Outdoor Settings}
\begin{figure}
    \centering
    \includegraphics[width=\linewidth, trim=0 200 0 0, clip]{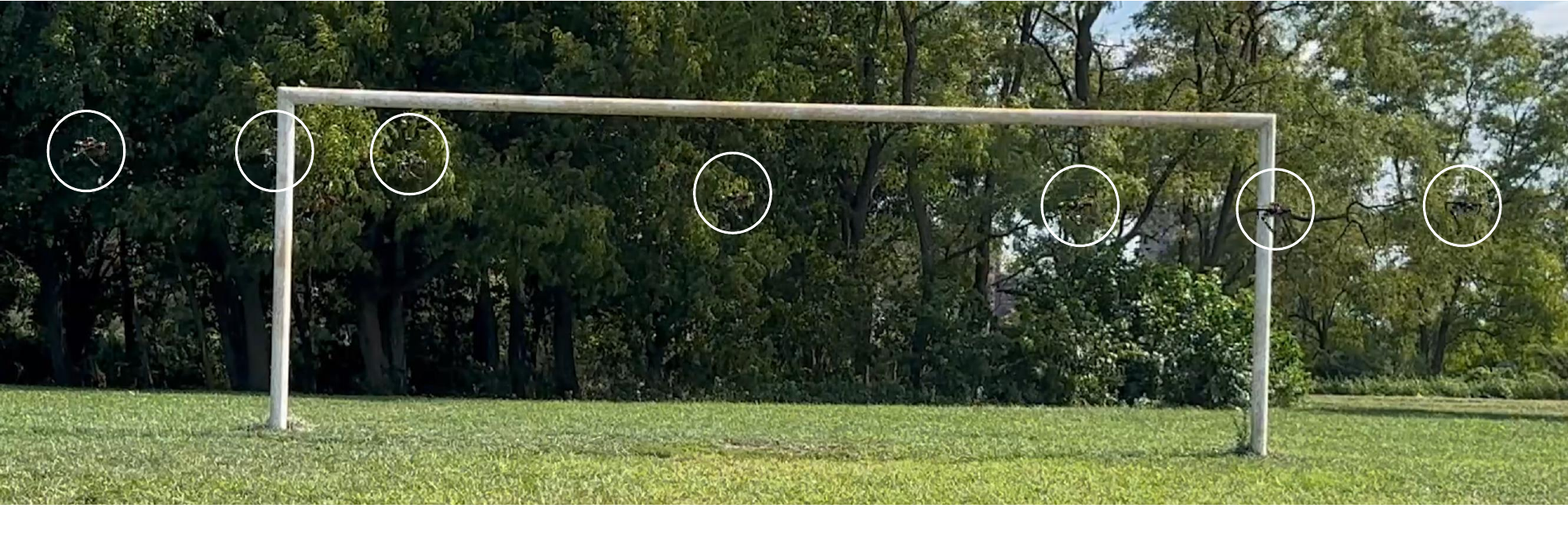}
    \includegraphics[width=\linewidth, trim=0 0 0 0, clip]{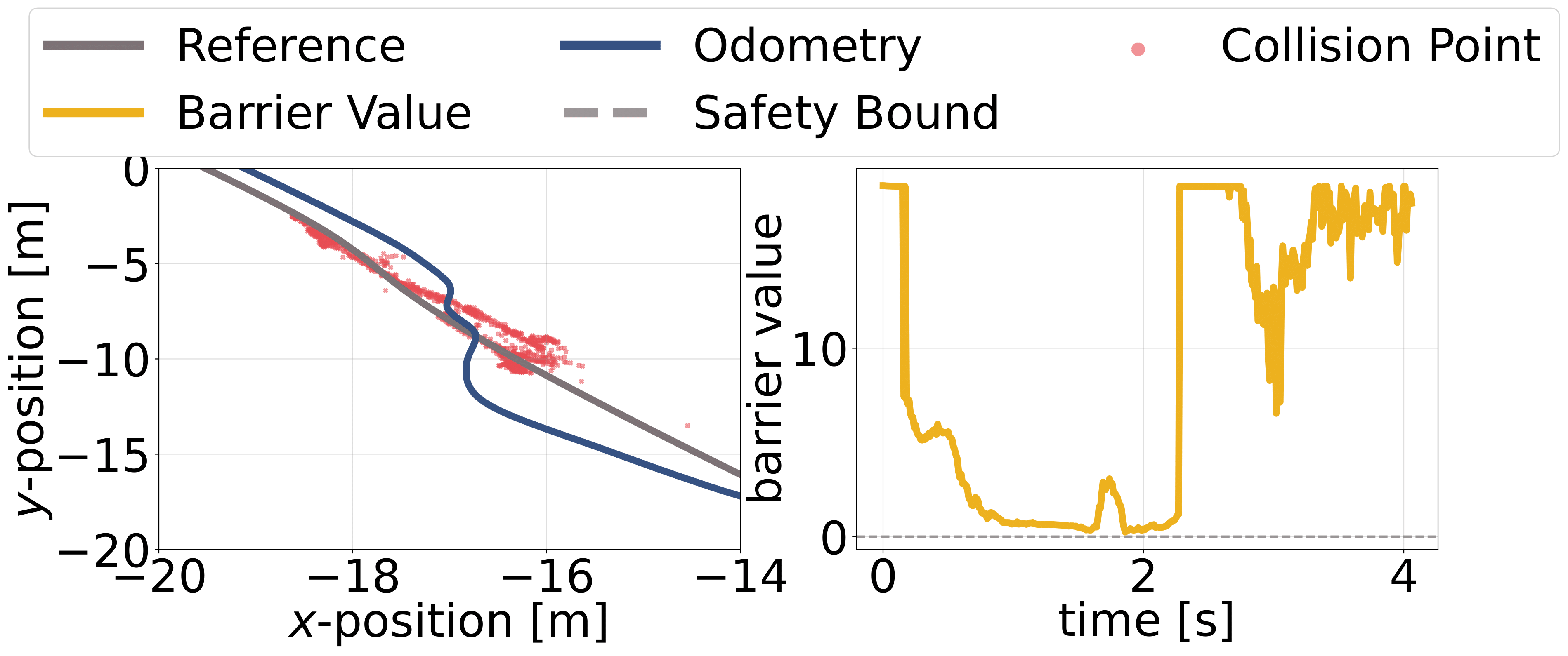}
    \caption{
    \textbf{Safe navigation through soccer goal's poles.} The quadrotor successfully navigates through narrow passages of a soccer goal, demonstrating its reactive collision avoidance capabilities in a real-world outdoor experiment.
    }
    \label{fig:soccer_goal_poles}
    \vspace{-1em}
\end{figure}

We evaluate the system’s performance in outdoor environments when navigating through forest obstacles and maneuvering around the poles of a soccer goal. These setups test the system's reactive collision avoidance capabilities, with the forest scenario presenting a cluttered environment and the soccer goal involving narrow passages and sudden obstacles. Results are shown in Figure~\ref{fig:initial_figure} and Figure~\ref{fig:soccer_goal_poles}.

The quadrotor demonstrates agile safe flight with speeds up to $7.2~\si{m/s}$, highlighting the system's robustness and adaptability in complex outdoor settings. Testing outdoors introduces challenges such as variable lighting, dynamic elements like wind, and inconsistent terrain that impact sensor performance and reliability. A critical aspect of these experiments is addressing the perception limitations of the Intel RealSense D435i, which measures depth up to $3~\si{\meter}$. By enhancing raw depth data with a monocular depth estimation network, the system extends its perception range, allowing for earlier obstacle detection and improved responsiveness.

\subsection{Analysis: Polynomial Order for Depth Completion} \label{subsec:analysis_poly_order}  
We analyze the impact of polynomial order on depth completion accuracy using NYUv2~\cite{silbermannyuv2} (indoor) and KITTI~\cite{Geiger2013IJRR} (outdoor). Ground truth depth maps serve as the reference, and polynomial functions of different orders are fitted to monocular depth estimates. Key metrics include Root Mean Square Error (RMSE), which emphasizes large depth discrepancies, and Absolute Relative Error (AbsRel), which captures proportional accuracy.

As shown in Table~\ref{tab:ablation_study_poly_order}, the quadratic model achieves the lowest errors on NYUv2, while the linear model performs better on KITTI in terms of RMSE and AbsRel, likely due to the nature of outdoor depth variations and sparser LiDAR-based ground truth. The cubic model tends to overfit in both cases. Additionally, KITTI exhibits higher RMSE and AbsRel values due to its greater depth range and sparser labels compared to the denser RGBD data in NYUv2.

\subsection{Analysis: Computational Complexity}
We analyze the computational complexity of our system by examining the trade-offs between image resolution and network inference speed, as well as the impact of the number of CBFs on NMPC run-time. The evaluation considers the mean performance across $10$ runs.

\begin{figure}[t]
    \centering
    \includegraphics[width=\linewidth, trim=0 0 0 0, clip]{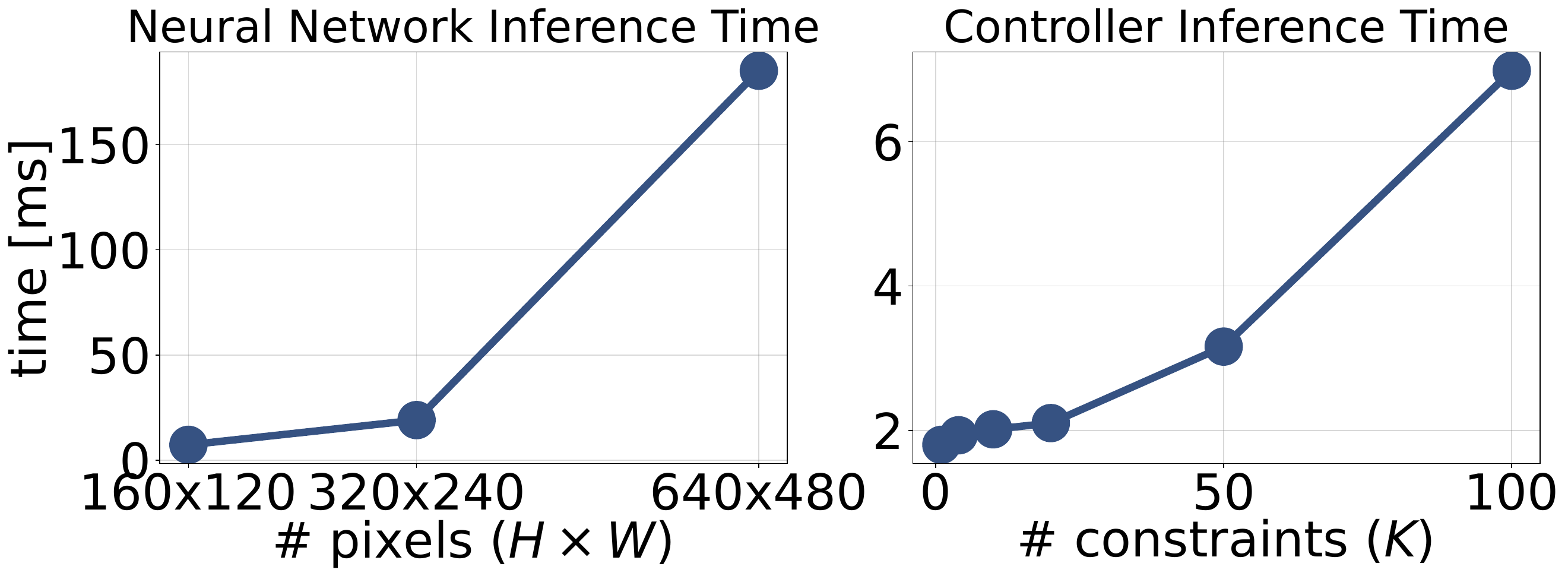}
    \caption{
        \textbf{Analysis on computational complexity.} 
        The left plot shows increased inference time with higher image resolution, highlighting the trade-off between speed and detail. The right plot demonstrates a moderate rise in controller run-time with more CBFs, maintaining feasibility.
    }
    \label{fig:ablation_study_comput_complexity}
    \vspace{-1em}
\end{figure}

Figure~\ref{fig:ablation_study_comput_complexity} highlights two critical findings. 
The left plot shows that increasing image resolution from $160 \times 120$ to $640 \times 480$ dramatically increases inference time, from $7.3~\si{ms}$ to $185.1~\si{ms}$. The steep rise in computation time between $320 \times 240$ and $640 \times 480$ led us to choose the former resolution for our system, balancing depth estimation accuracy and real-time performance.
The right plot demonstrates that the NMPC run-time scales efficiently with the number of CBFs. As the number of constraints increases from $1$ to $100$, the run-time only grows from $1.8~\si{ms}$ to $7.0~\si{ms}$, indicating that the CBFs do not significantly impact computation time. This confirms the feasibility and real-time applicability of our approach, even with multiple constraints.

\begin{table}[t]
    \addtolength{\tabcolsep}{-0.54em}
    \centering
\caption{
    \textbf{Analysis on polynomial order for depth completion.}
    Performance of different polynomial orders (linear, quadratic, and cubic) used for depth completion on different datasets. The quadratic model achieves the best accuracy on NYUv2, while the linear model performs slightly better on KITTI in terms of RMSE and AbsRel.
    \label{tab:ablation_study_poly_order}
}
    \begin{tabular}{llccccc}
        \toprule\toprule
        \textbf{Polyn.} & \textbf{Polyn.} &
        \multicolumn{2}{c}{\textbf{NYUv2}~\cite{silbermannyuv2}} && \multicolumn{2}{c}{\textbf{KITTI}~\cite{Geiger2013IJRR}}\\
        \cline{3-4} \cline{6-7}
        \textbf{Order} & \textbf{Equation} &
        \textbf{RMSE} & \textbf{AbsRel} &&
        \textbf{RMSE} & \textbf{AbsRel} \\
        \midrule
        Linear & $y=aX+b$ & 
        2.43 & 0.73 &&
        5.11 & 0.13\\
        Quadratic & $y=aX^2+bX+c$ & 
        0.39 & 0.08 &&
        5.25 & 0.13\\
        Cubic & $y=aX^3+bX^2+cX+d$ &
        3.01 & 0.81 &&
        5.37 & 0.13\\
        \bottomrule\bottomrule
    \end{tabular}
\end{table}

\section{Conclusion and Future Directions} \label{sec:conclusion}
This paper presented a new obstacle avoidance framework that integrates perception constraints directly into an NMPC, enabling agile and safe navigation for autonomous robots. By tightly addressing perception, planning, and control, the proposed approach effectively navigated complex environments in real-time, as demonstrated through extensive indoor and outdoor experiments. The system’s ability to dynamically adjust control barrier functions ensured robust performance without reliance on specific environmental knowledge, making it suitable for various applications, including target tracking and autonomous exploration.

Future work will extend the system to handle dynamic obstacles by incorporating obstacle kinematics prediction into NMPC. Additionally, transitioning to vision-based navigation will allow the quadrotor to rely solely on camera inputs, reducing hardware complexity and enhancing applicability to compact scenarios.

\clearpage

\bibliographystyle{IEEEtran}
\bibliography{references}

\end{document}